\documentclass[letterpaper, 10 pt, conference]{ieeeconf}
\usepackage{graphicx}
\usepackage{courier}
\usepackage{cite}
\usepackage{bm}
\usepackage{nccmath}
\usepackage{amssymb}
\usepackage{mathtools}
\usepackage{amsmath}
\usepackage{indentfirst}
\usepackage{algorithm}
\usepackage[pagebackref=true, colorlinks=true, linkcolor=red, citecolor=blue, urlcolor=blue]{hyperref}

\usepackage{algpseudocode}
\usepackage{orcidlink}

\DeclareMathOperator*{\argmin}{argmin}

\DeclarePairedDelimiter{\nint}\lfloor\rceil

\usepackage{colortbl}
\usepackage{multirow}
\usepackage{array}

\makeatletter
\newcommand{\mathleft}{\@fleqntrue\@mathmargin0pt}
\newcommand{\mathcenter}{\@fleqnfalse}
\makeatother

\algnewcommand\algorithmicnot{\textbf{not}}
\algdef{SE}[IF]{IfNot}{EndIf}[1]{\algorithmicif\ \algorithmicnot\ #1\ \algorithmicthen}{\algorithmicend\ \algorithmicif}%
\algnewcommand\algorithmicforeach{\textbf{for each}}
\algdef{S}[FOR]{ForEach}[1]{\algorithmicforeach\ #1\ \algorithmicdo}
\IEEEoverridecommandlockouts
\begin{document}

\title{\textbf{KnotDLO: Toward Interpretable Knot Tying}}

\author{
Holly Dinkel\textsuperscript{1,3}\orcidlink{0000-0002-7510-2066}, 
Raghavendra Navaratna\textsuperscript{1}\orcidlink{0009-0000-7567-0955},
Jingyi Xiang\textsuperscript{2}\orcidlink{0000-0003-0727-3098},
Brian Coltin\textsuperscript{3}\orcidlink{0000-0003-2228-6815},
Trey Smith\textsuperscript{3}\orcidlink{0000-0001-8650-8566},
Timothy Bretl\textsuperscript{1}\orcidlink{0000-0001-7883-7300}
\thanks{
\textsuperscript{1}Department of Aerospace Engineering and Coordinated Science Laboratory, University of Illinois at Urbana-Champaign, Urbana, IL, 61801. \texttt{\{hdinkel2, rsn3, tbretl\}@illinois.edu.}}
\thanks{\textsuperscript{2}Department of Electrical Engineering and Coordinated Science Laboratory, University of Illinois at Urbana-Champaign, Urbana, IL, 61801. \texttt{jingyix4@illinois.edu.}}
\thanks{\textsuperscript{3}Intelligent Robotics Group, NASA Ames Research Center, Moffett Field, CA, 94035 USA \texttt{\{brian.coltin, trey.smith\}@nasa.gov.}}
}

\maketitle

\begin{abstract} 

This work presents KnotDLO, a method for one-handed Deformable Linear Object (DLO) knot tying that is robust to occlusion, repeatable for varying rope initial configurations, interpretable for generating motion policies, and requires no human demonstrations or training. Grasp and target waypoints for future DLO states are planned from the current DLO shape. Grasp poses are computed from indexing the tracked piecewise linear curve representing the DLO state based on the current curve shape and are piecewise continuous. KnotDLO computes intermediate waypoints from the geometry of the current DLO state and the desired next state. The system decouples visual reasoning from control. In 16 trials of knot tying, KnotDLO achieves a 50\% success rate in tying an overhand knot from previously unseen configurations. 

\end{abstract}
\IEEEpeerreviewmaketitle

\section{Introduction}
\label{sec: introduction}

Manipulating Deformable Linear Objects (DLOs) into knots is useful for many robotic tasks, including climbing~\cite{page2023tether, specian2015friction}, surgery~\cite{mayer2006heart,wang2008novel,wang2010suturing,chow2013improved}, household robotics~\cite{viswanath2021disentangling}, and industrial robotics~\cite{mitrano2021learning,li2018vision,jin2021trajectory}. Manipulating DLOs can be categorized based on the type of DLO shape information used as input. Tasks such as insertion~\cite{wang2015online}, disentangling~\cite{wakamatsu2006knotting,viswanath2021disentangling}, and clearing DLOs from the path to other objects~\cite{rojas2022autonomous} may only require information such as the location of the tips, the locations of crossings, or the locations of inner and external knot segments rather than a full shape estimate. Tasks such as winding, braiding, and knotting may benefit from a complete shape estimate~\cite{xiang2023multidlo}. One challenge is representing the DLO topology, especially when the DLO is occluded by itself or the environment as it is manipulated. Recent work in real-time visual DLO tracking enables occlusion-robust topology representation for autonomous knot tying. KnotDLO achieves repeatable knot tying using topological waypoints validated in overhand knot tying experiments with a success rate of 50\%, compared to the success rate of the learning-based state-of-the-art method of 66\%~\cite{sundaresan2020learning}. The topological waypoints enable transformation between knot states using movement primitives independent of task parameterizations such as initial DLO configuration or DLO length. KnotDLO computes geometric grasp and target poses from the geometry of the DLO and are a subset of a continuous space of grasp and target poses defined by the DLO topology.

\begin{figure}
\centering
\includegraphics[width=
\columnwidth]{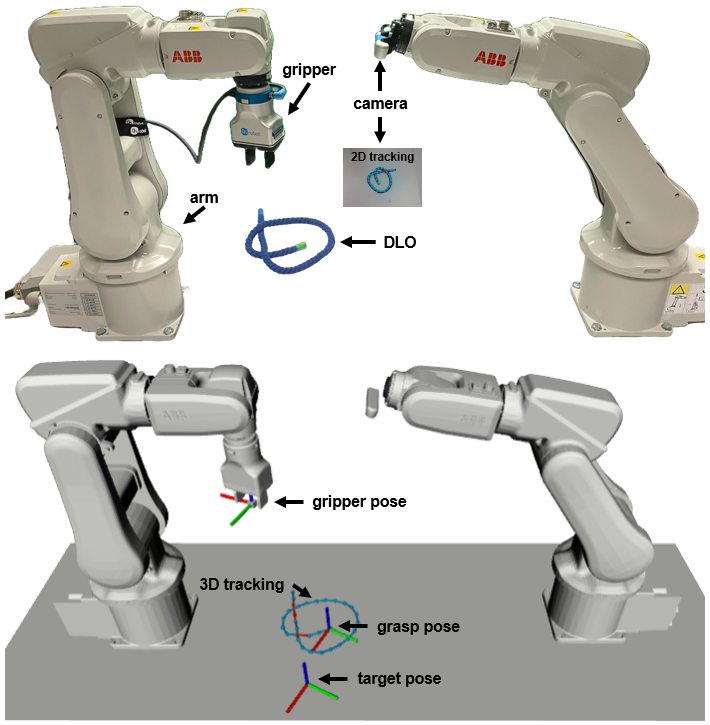}
\caption{One robot arm with a parallel gripper grasps the DLO and moves it between topological waypoints. The DLO topology is tracked from imagery collected by a camera mounted on the last link of a second robot arm.\vspace{-1.5em}}
\label{fig: scene}
\end{figure}

\begin{figure*}
\centering
\includegraphics[width=\textwidth]{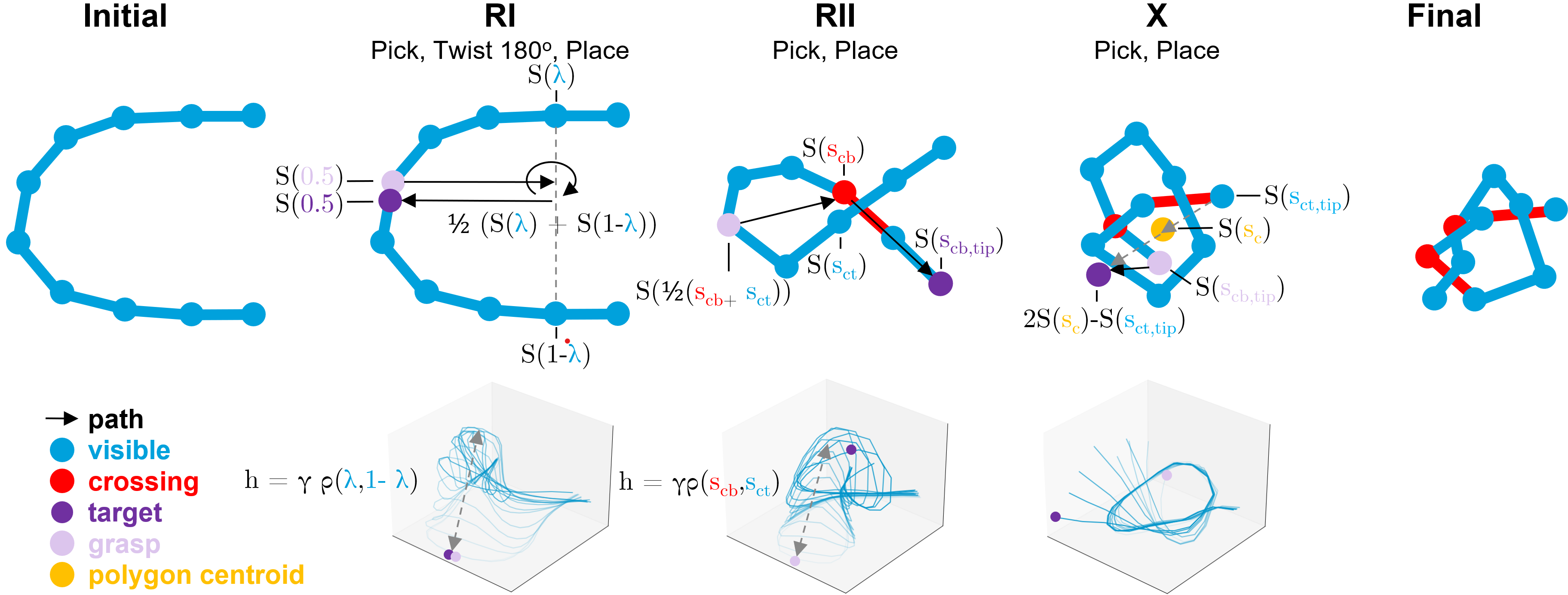}
\caption{Given an initial symmetric, curve-shaped DLO configuration, the planning system computes grasp and target indices from the topology to compose a sequence knot tying movement primitives. (Top) Sequentially performing Reidemeister Move I (RI), Reidemeister Move II (RII), and the Cross Move (X) results in an overhand knot. (Bottom) The tracking system tracks the shape of the DLO as it moves. \vspace{-1.5em}}
\label{fig: procedure}
\end{figure*}

\begin{figure}
\centering
\includegraphics[width=0.7\columnwidth]{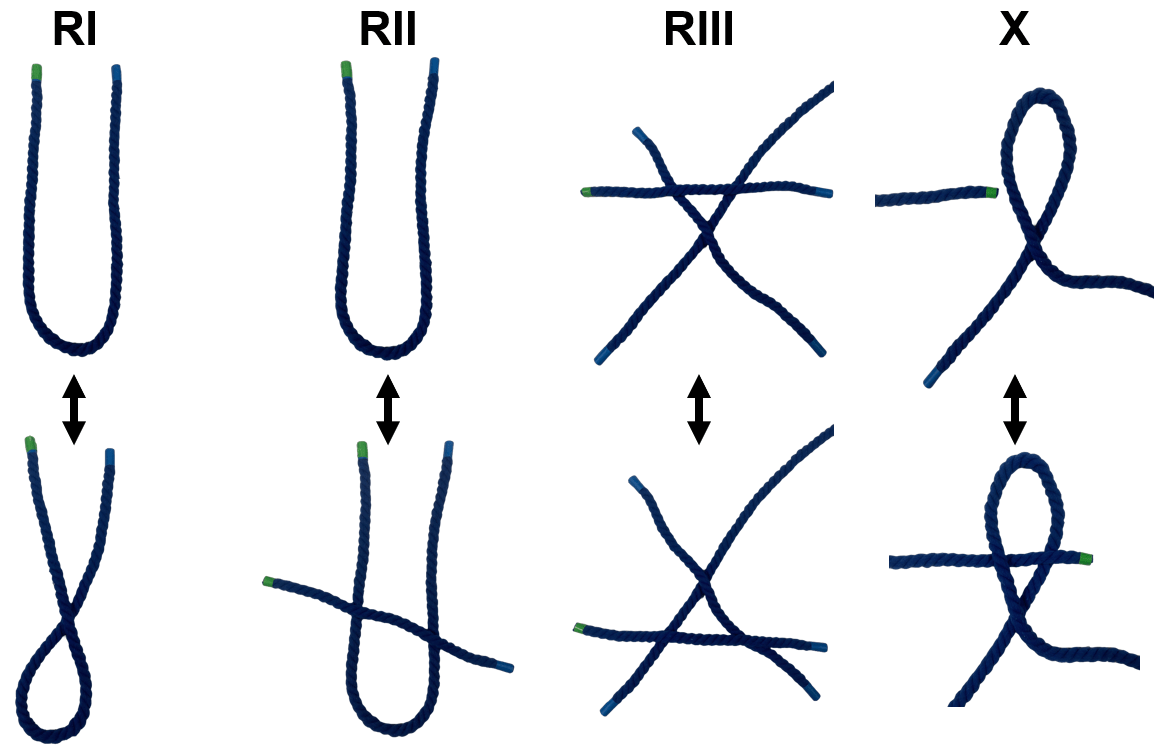}
\caption{Four movement primitives can transform a DLO between any knot states. (Left) Reidemeister Move I twists the DLO to add or remove one crossing. (Center left) Reidemeister Move II slides one strand over one loop to add or remove two crossings. (Center right) Reidemeister Move III slides one strand over a crossing to transform the knot state around the crossing. (Right) The Cross Move pokes the tip of a DLO through a loop to add or remove two crossings. \vspace{-1.5em}}
\label{fig: reidemeister}
\end{figure}

\section{Related Work}
\label{sec: related-work}

The availability of an accurate DLO topology for robotic knot tying has conventionally been limited by perception systems which struggle to follow the shape of the DLO as it is occluded within the environment or as it is occluded by itself during the formation of intersections. One method overcomes this by learning a fast and differentiable neural network model of the DLO dynamics to augment segmentation-based shape estimation with physics~\cite{yan2020supervised,yan2020TMP}. Another method learns DLO configuration-independent dense depth object descriptors from synthetic ropes rendered in Blender and uses these descriptors to learn manipulation action sequences with visual imitation of demonstrations~\cite{sundaresan2020learning}. This method encodes geometric structure on the descriptors and learns them in simulation without any robot motion learning. This accelerates training time substantially from work which requires collecting robot motion demonstration data for imitation~\cite{pathak2018zeroshot, nair2017combining} and enables transferring the learned visual representation across domains. Other work performs DLO shape manipulation with visual feedback for shape control without self-occlusion \cite{yan2020TMP,yu2022shapecontrol,yu2023modellearning}. Recent advances in DLO perception enable DLO shape tracking under environment occlusion~\cite{caporali2022ariadneplus,wang2021cdcpd2,xiang2023trackdlo}. These methods also accurately track shape under self-occlusion, offering new opportunities for full DLO state representation for knot tying.

The Reidemeister moves are a set of movement primitives which add or remove links in a knot projection, creating a new projection. Reidemeister Move I (RI) adds or removes one crossing through twisting or untwisting. Reidemeister Move II (RII) adds or removes two crossings by sliding one straight segment over one curved segment. Reidemeister Move III (RIII) transforms the topology of a crossing by sliding one segment over or under a crossing \cite{reidemeister1932knottheory}. The Cross Move (X) was introduced for tying objects which are not closed loops in the unknotted state. It passes the tip of the rope through a loop in the knot to create a crossing. Composing the RI, RII, and X moves in sequence results in an overhand knot.

\section{One-Handed Overhand Knot Manipulation}

This work presents an overhand knot tying system using real-time visual DLO shape tracking to compute manipulation waypoints~\cite{xiang2023trackdlo}. The topology of a DLO is represented as a piecewise linear curve, $S$, constructed from a set of $M$ control points, $C^{M\times 3} \in \mathbb{R}^3$. The curve $S$ is parameterized at time $t$ by curvilinear length, $s^t$, and total length, $L$, with

\small
\begin{equation}
    S: s \in [0, 1] \rightarrow S(s^t;L) \in \mathbb{R}^3.
\end{equation}
\normalsize

\noindent The head and tail positions of the DLO are $S(0)$ and $S(1)$. Two curvilinear lengths, $s_i^t$ and $s_j^t$, are determined to be on the same segment of $S(s^t;L)$ if $\left| \text{idx}_i - \text{idx}_j \right| \leq 1$, where

\small
\begin{equation}
    \text{idx}_i = \nint*{s_i \times (M-1)}.
\end{equation}
\normalsize

\noindent The geodesic distance defines the length of the shortest path along a manifold. The geodesic distance, $\rho(s_i^t,s_j^t)$, is

\small
\begin{fleqn}
    $\rho(s_i^t,s_j^t) =$
\end{fleqn}
\begin{equation}
\label{eq: geodesic-s-to-s}
    \begin{cases} 
        \left| s_j^t - s_i^t \right| \cdot \left\| S(s_j^t) - S(s_i^t) \right\| \text{ if } \left| \text{idx}_j - \text{idx}_i \right| \leq 1 \\
        \left| 1 - s_j^t \right| \cdot \left\| S(s_i^t) - C_{\text{end,i}}^t \right\| + \left| s_j^t \right| \cdot \left\| S(s_j^t) - C_{\text{end,j}}^t \right\| \\ 
        + \sum_{m = \text{end,i} + 1}^{\text{end,j}} \left\| C_m - C_{m-1} \right\| \text{ otherwise}
    \end{cases},
\end{equation}
\normalsize

\noindent where $\| \cdot \|$ is the $l_2$ norm and

\small
\begin{equation}
\vspace{-0.1em}
    C_{\text{end,i}}^t = \argmin_{m \in M} \| S(s_i^t) - C_m \|.
\vspace{-0.1em}
\end{equation}
\normalsize

\begin{figure}
\centering
\includegraphics[width=0.7\columnwidth]{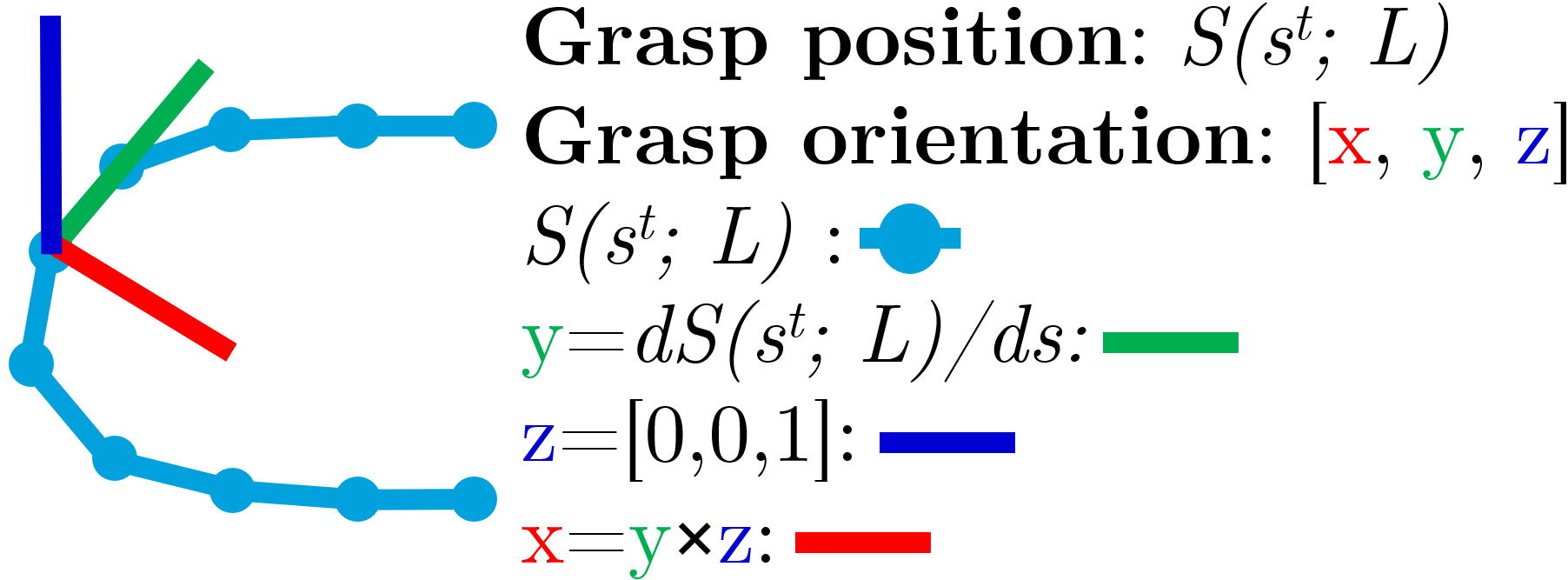}
\caption{A grasp pose can be computed given the curve representation of the tracking result, $S(s^t; L)$, and the curvilinear length at a given time step, $s^t$. The grasp position is $S(s^t)$ and the orientation is computed from the unit-length tangent vector to the curve, $dS(s^t;L)/ds$ assuming semi-planar knot tying with $z=\left[0,0,1\right]$. \vspace{-1.5em}}
\label{fig: grasp_pose}
\end{figure}

The DLO topology is used to plan grasp and target poses for a robot arm with a gripper to tie an overhand knot. As shown in Fig.~\ref{fig: grasp_pose}, the orientation of these poses are computed using the semi-planar manipulation constraint (i.e., $z=\left[0,0,1\right]$), where the $y-$axis of the transform is the unit-length tangent vector to the curve at the desired point, $dS(s^t;L)/ds$, and the $x-$axis is the axis normal to the $y-z$ plane. Waypoints transforming each grasp pose to the target pose are designed to reduce effects from sliding on the manipulation plane. Seven curvilinear lengths---$s=0.5$, $s=\lambda$, $s=1-\lambda$, $s_{\text{cb}}$ (the crossing bottom), $s_{\text{ct}}$ (the crossing top), $s_{\text{cb, tip}}$ (the undertip), and $s_{\text{ct, tip}}$ (the overtip)---are selected to tie an overhand knot for a DLO in an initial symmetric, curved configuration as shown in Fig.~\ref{fig: procedure}. The DLO is first grasped at $S(0.5)$, lifted to a height $h = \gamma \rho(\lambda,1-\lambda)$, and translated to the $x-y$ positions given by the midpoint between the two parallel lengths, $\frac{1}{2}\left(S(\lambda) + S(1-\lambda)\right)$. At this waypoint, the grasped point is rotated by $\frac{\pi}{2}$ radians about the $z$-axis, then returned to the start position of the move to complete RI. The top crossing, $s_{\text{ct}}$, is calculated as the control point closest in Euclidean distance to the self-occluded control point estimated from the tracking algorithm using their 2D projections in pixel space~\cite{xiang2023trackdlo}. The undertip is computed as the tip closest to the crossing bottom using Eq. (\ref{eq: geodesic-s-to-s}). To perform RII, the DLO is grasped at $S(0.5)$, lifted to a height $h = \gamma \rho(s_{\text{cb}},s_{\text{ct}})$, translated to $S(s_{\text{cb}})$, translated to $S(s_{\text{cb,tip}})$, and placed. Tracking often fails to accurately initialize the crossing topology after this step due to its sensitivity to depth resolution, however the X move requires grasping near $S(s_{\text{cb, tip}})$. Instead of computing $S(s_{\text{cb, tip}})$ from topology, $S(s_{\text{cb, tip}})$ is computed as the tip nearest to the centroid of a convex polygon formed from the tracked points along the DLO~\cite{gillies2022shapely}. To improve robustness, the closest node to the polygon centroid within a radius of $r$ nodes from the undertip is selected as the grasp point. The target position for the X move is computed as $S(s_c) + \left(S(s_c) - S(s_{ct,tip})\right)$ and the target orientation is the same as the grasp orientation.

\begin{table}
\centering
\caption{Knot Tying Success Rate By Move}
\begin{tabular}{|c|c|c|c|}
    \hline
    RI & RII & X & \textbf{Total} \\
    \hline
    0.937 & 0.867 & 0.615 & \textbf{0.500} \\ 
    \hline
\end{tabular}
\label{tab: per-move-success-rate}
\vspace{-2em}
\end{table}

\section{Experiments}

\begin{figure}
\centering
\includegraphics[width=\columnwidth]{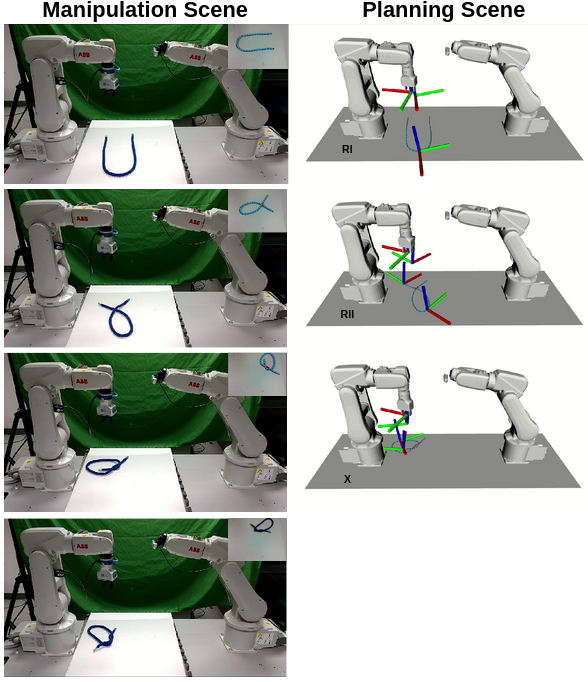}
\caption{A small number of waypoints designed based on the rope topology are required to tie an untightened overhand knot.}
\label{fig: planning}
\end{figure}

The DLO is assumed to start in a symmetric curved shape and able to be distinguished from the background using image-based thresholding in Hue, Saturation, and Value (HSV) color space. One tip of the rope is covered with green tape to break object symmetry when initializing the object shape. In 16 experiments analyzing the repeatability of one-handed knot manipulation, the system achieves a 50\% success rate in tying an an overhand knot from previously unseen initial DLO configurations. All experiments used an Intel RealSense d435 camera for vision with default tracking parameters in TrackDLO with the number of nodes used as control points, $M$, as $M=30$. The DLO is a blue rope with $L = 0.88 \mathrm{m}$, $\lambda=0.1$, $\gamma=0.4$ and $r=5$. The manipulation system includes one ABB IRB120 manipulator with an OnRobot 2FG7 gripper. Manipulation assumes the perceived topology of the DLO is accurate before performing RI and RII, but not before performing RIII. Figure~\ref{fig: planning} shows the steps of knot tying in the manipulation scene, which shows the transformation between knot states, and the planning scene, which shows the poses of waypoints. Table \ref{tab: per-move-success-rate} reports per-move knot tying success rates. Table \ref{tab: comparison} reports knot tying success rates from the literature.

\begin{table}
\centering
\caption{Overall Overhand Knot Tying Success Rates}
\begin{tabular}{|c|c|c|c|}
    \hline
    \textbf{KnotDLO (Ours)} & DDOD~\cite{sundaresan2020learning} & GSP~\cite{pathak2018zeroshot} & Imitation~\cite{nair2017combining} \\
    \hline
    0.50 & \textbf{0.66} & 0.6 & 0.38 \\
    \hline
\end{tabular}
\label{tab: comparison}
\end{table}

\section{Conclusions}
\label{conclusions}

This work introduced KnotDLO, a system for semi-planar one-handed overhand knot tying. KnotDLO is the first knot tying method using topological DLO state tracking from a perception system as input. The knot tying system was evaluated in repeated knot tying trials from different initial symmetric curved configurations. The key advantages of this system are its interpretability and robustness to occlusion, including its ability to resolve points of crossing in the DLO. This method shows promise for further development and deployment as an explainable system for knot tying. In future work, KnotDLO could also be used to automatically collect high-quality demonstration data to learn multi-step manipulation policies comprising sequences of pick-and-place actions~\cite{seita2021deformable}.

\section*{Acknowledgements}

\noindent \small The authors thank the members of the Representing and Manipulating Deformable Linear Objects project (\href{https://github.com/RMDLO}{https://github.com/RMDLO}) and the teams developing the open-source software used in this project \cite{ros, opencv_library, harris2020numpy, hunter2007matplotlib, zhou2018open3d, virtanen2020scipy, koide2019handeye, eigenweb, pcl, gillies2022shapely}. This work was supported by the NASA Space Technology Graduate Research Opportunity 80NSSC21K1292, the P.E.O. Scholar Award, and the Zonta International Amelia Earhart Fellowship.

\bibliographystyle{IEEEtran}
\bibliography{abbreviations,references}
\end{document}